\documentclass[10pt,twocolumn,letterpaper]{article}

\usepackage{times}
\usepackage{epsfig}
\usepackage{graphicx}
\usepackage{amsmath}
\usepackage{amssymb}
\usepackage{indentfirst}

\begin{document}

\title{FAT: Fast Adjustable Threshold for Uniform Neural Network Quantization (Winning Solution on LPIRC-II)}

\author{Alexander Goncharenko\\
Expasoft LLC, Novosibirsk, Russia\\
Novosibirsk State University, Novosibirsk, Russia\\
{\tt\small a.goncharenko@expasoft.ru}
\and
Sergey Alyamkin\\
Expasoft LLC, Novosibirsk, Russia\\
{\tt\small s.alyamkin@expasoft.ru}
\and
Andrey Denisov\\
Expasoft LLC, Novosibirsk, Russia\\
Novosibirsk State University, Novosibirsk, Russia\\
{\tt\small a.denisov@expasoft.ru}
\and
Evgeny Terentev\\
Microtech, Moscow, Russia        \\
{\tt\small et@microtech.ai}
}

\date{} 
\maketitle

\begin{abstract}
The neural network quantization is highly desired procedure to perform before running neural networks on mobile devices. Quantization without fine-tuning leads to accuracy drop of the model, whereas commonly used training with quantization is done on the full set of the labeled data and therefore is both time- and resource-consuming. Real life applications require simplification and acceleration of quantization procedure that will maintain the accuracy of full-precision neural network, especially for modern mobile neural network architectures like Mobilenet-v1, MobileNet-v2 and MNAS.

Here we present two methods to significantly optimize the training with quantization procedure. The first one is introducing the trained scale factors for discretization thresholds that are separate for each filter. The second one is based on mutual rescaling of consequent depth-wise separable convolution and convolution layers. Using the proposed techniques, we quantize the modern mobile architectures of neural networks with the set of train data of only $\sim$ 10\% of the total ImageNet 2012 sample. Such reduction of train dataset size and small number of trainable parameters allow to fine-tune the network for several hours while maintaining the high accuracy of quantized model (accuracy drop was less than 0.5\%). Ready-for-use models and code are available at: https://github.com/agoncharenko1992/FAT-fast-adjustable-threshold.

\end{abstract}

\section{Introduction}
\label{Introduction}
Mobile neural network architectures \cite{1,2,3} allow running AI solutions on mobile devices due to the small size of models, low memory consumption, and high processing speed while providing a relatively high level of accuracy in image recognition tasks. In spite of their high computational efficiency, these networks continuously undergo further optimization to meet the requirements of edge devices. One of the promising optimization directions is to use quantization to int8, which is natively supported by mobile processors, either with or without training. Both methods have certain advantages and disadvantages.

Quantization of the neural network without training is a fast process as in this case a pre-trained model is used. However, the accuracy of the resultant network is particularly low compared to the one typically obtained in commonly used mobile architectures of neural networks \cite{4}. On the other hand, quantization with training is a resource-intensive task which results in low applicability of this approach. 

Current article suggests a method which allows speeding up the procedure of training with quantization and at the same time preserves a high accuracy of results for 8-bit discretization.

\section{Related work}
\label{Related}

In general case the procedure of neural network quantization implies discretization of weights and input values of each layer. Mapping from the space of float32 values to the space of signed integer values with $n$ significant digits is defined by the following formulae: 

\begin{equation}
S_w = \frac{2^n - 1}{T_w}
\end{equation}

\begin{equation}
T_w = max |W| 
\end{equation}

\begin{equation}
W_{int} = \lfloor {S_w \cdot W} \rceil
\end{equation}

\begin{equation} 
\begin{split}
& W_q = clip(W_{int}, -(2^{n-1} - 1), 2^{n-1} - 1) = \\
& = min(max(W_{int}, -(2^{n-1} - 1)), 2^{n-1} - 1) 
\end{split}
\end{equation}

Here $\lfloor \, \rceil$ is rounding to the nearest integer number, $W$ – weights of some layer of neural network, $T$ – quantization threshold, $max$ calculates the maximum value across all axes of the tensor. Input values can be quantized both to signed and unsigned integer numbers depending on the activation function on the previous layer.  

\begin{equation}
S_i = \frac{2^n - 1}{T_i}  
\end{equation}

\begin{equation}
T_i = max |I|  
\end{equation}

\begin{equation}
I_{int} = \lfloor {S_i \cdot I} \rceil
\end{equation}

\begin{equation}
I^{signed}_q = clip(I_{int}, -(2^{n-1} - 1), 2^{n-1} - 1) 
\end{equation}

\begin{equation}
I^{unsigned}_q = clip(I_{int}, 0, 2^n - 1) 
\end{equation}

After all inputs and weights of the neural network are quantized, the procedure of convolution is performed in a usual way. It is necessary to mention that the result of operation must be in higher bit capacity than operands. For example, in Ref. \cite{5} authors use a scheme where weights and activations are quantized to 8-bits while accumulators are 32-bit values.

Potentially quantization threshold can be calculated on the fly, which, however, can significantly slow down the processing speed on a device with low system resources. It is one of the reasons why quantization thresholds are usually calculated beforehand in calibration procedure. A set of data is provided to the network input to find desired thresholds (in the example above - the maximum absolute value) of each layer. Calibration dataset contains the most typical data for the certain network and this data does not have to be labeled according to procedure described above.

\subsection{Quantization with knowledge distillation}

Knowledge distillation method was proposed by G. Hinton \cite{6} as an approach to neural network quality improvement. Its main idea is training of neural networks with the help of pre-trained network. In Refs. \cite{7,8} this method was successfully used in the following form: a full-precision model was used as a model-teacher, and quantized neural network - as a model-student. Such paradigm of learning gives not only a higher quality of the quantized network inference, but also allows reducing the bit capacity of quantized data while keeping an acceptable level of accuracy.

\subsection{Quantization without fine-tuning}

Some frameworks allow using the quantization of neural networks without fine-tuning. The most known examples are TensorRT \cite{22}, Tensorflow \cite{9} and Distiller framework from Nervana Systems \cite{23}. However, in the last two models calculation of quantization coefficients is done on the fly, which can potentially slow down the operation speed of neural networks on mobile devices. In addition, to the best of our knowledge, TensorRT framework does not support quantization of neural networks with the architectures like MobileNet.

\subsection{Quantization with training / fine-tuning}

One of the main focus points of research publications over the last years is the development of methods that allow to minimize the accuracy drop after neural network quantization. The first results in this field were obtained in Refs. \cite{10,11,12,13}. The authors used the Straight Through Estimator (STE) \cite{14} for training the weights of neural networks into 2 or 3 bit integer representation. Nevertheless, such networks had substantially lower accuracy than their full-precision analogs. 

The most recent achievements in this field are presented in Refs. \cite{15,16} where the quality of trained models is almost the same as for original architectures. Moreover, in Ref. \cite{16} the authors emphasize the importance of the quantized networks ensembling which can potentially be used for binary quantized networks. In Ref. \cite{5} authors report the whole framework for modification of network architecture allowing further launch of learned quantized models on mobile devices. 

In Ref. \cite{17} the authors use the procedure of threshold training which is similar to the method suggested in our work. However, the reported approach has substantial shortcomings and cannot be used for fast conversion of pre-trained neural networks on mobile devices. First of all it has a requirement to train threshold on the full ImageNet dataset \cite{18}. Besides, it has no examples demonstrating the accuracy of networks used as standards for mobile platforms. 

In current paper we propose a novel approach to set the quantization threshold with fast fine-tuning procedure on a small set of unlabeled data that allows to overcome the main drawbacks of known methods. We demonstrate performance of our approach on modern mobile neural network architectures (MobileNet-v2, MNAS).

\section{Method description}
\label{Method}

Under certain conditions (see Figures 1-2) the processed model can significantly degrade during the quantization process. The presence of outliers for weights distribution shown in Figure 1 forces to choose a high value for thresholds that leads to accuracy degradation of quantized model. 

Outliers can appear due to several reasons, namely specific features of calibration dataset such as class imbalance or non-typical input data. They also can be a natural feature of the neural network, that are, for example, weight outliers formed during training or reaction of some neurons on features with the maximum value. 

\begin{figure}[t]
\begin{center}
\includegraphics[width=\linewidth]{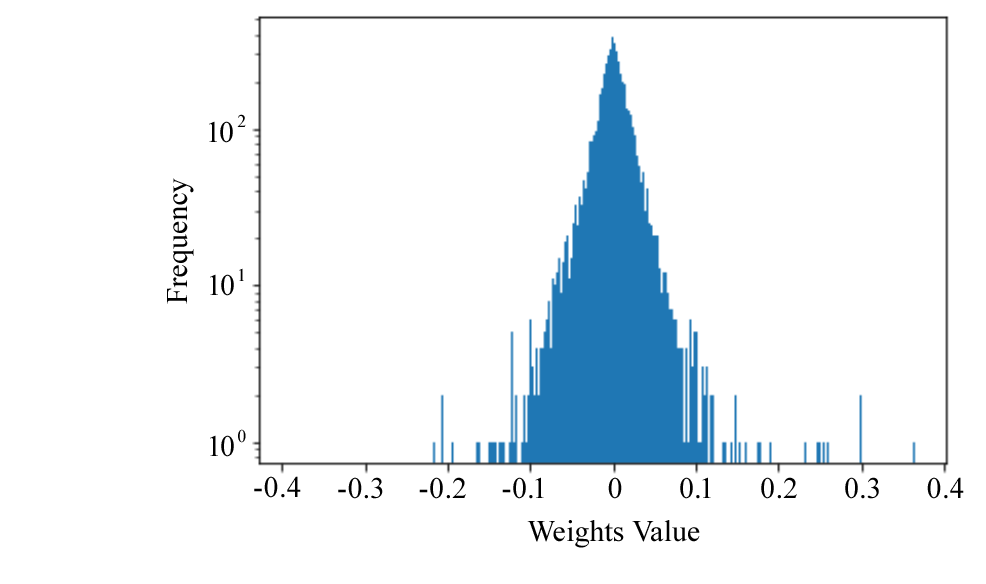}
\end{center}
\caption{Distribution of weights of ResNet-50 neural network before the quantization procedure.}
\label{weights-dist-1}
\end{figure}

\begin{figure}[t]
\begin{center}
\includegraphics[width=\linewidth]{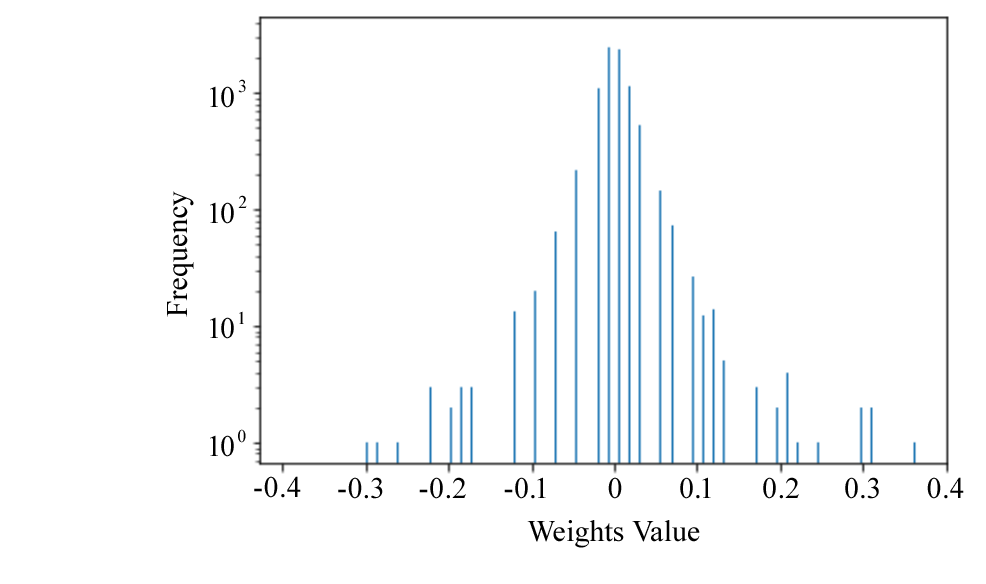}
\end{center}
\caption{Distribution of weights of ResNet-50 neural network after the quantization procedure. The number of values appeared in bins near zero increased significantly.}
\label{weights-dist-2}
\end{figure}

Overall it is impossible to avoid outliers completely because they are closely associated with the fundamental features of neural networks. However, it is possible to find a compromise between the value of threshold 
and distortion of other values during quantization and get a better quality of the quantized neural network.

\subsection{Quantization with threshold fine-tuning}
\subsubsection{Differentiable quantization threshold.} 

In Refs. \cite{11,13,14} it is shown that the Straight Through Estimator (STE) can be used to define a derivative of a function which is non-differentiable in the usual sense ($round$, $sign$, $clip$, etc). Therefore, the value which is an argument of this function becomes differentiable and can be trained with the method of steepest descent, also called the gradient descent method. Such variable is a quantization threshold and its training can directly lead to the optimal quality of the quantized network. This approach can be further optimized through some modifications as described below. 

\subsubsection{Batch normalization folding.} 

Batch normalization (BN) layers play an important role in training of neural networks because they speed up train procedure convergence \cite{19}. Before making quantization of neural network weights, we suggest to perform batch normalization folding with the network weights similar to method described in Ref. \cite{5}. As a result we obtain the new weights calculated by the following formulae:

\begin{equation}
W_{fold} = \frac{\gamma W}{\sqrt{\sigma^2 + \varepsilon}}
\end{equation}

\begin{equation}
b_{fold} = \beta - \frac{\gamma \mu}{\sqrt{\sigma^2 + \varepsilon}}
\end{equation}

We apply quantization to weights which were fused with the BN layers because it simplifies discretization and speeds up the neural network inference. Further in this article the folded weights will be implied (unless specified otherwise).

\subsubsection{Threshold scale.} 

All network parameters except quantization thresholds are fixed. The initial value of thresholds for activations is the value calculated during calibration. For weights it is the maximum absolute value. Quantization threshold $T$ is calculated as

\begin{equation}
T = clip(\alpha, min_{\alpha}, max_{\alpha})\cdot T_{max}
\end{equation}

\noindent where $\alpha$ is a trained parameter which takes values from $min_{\alpha}$ to $max_{\alpha}$ with saturation. The typical values of these parameters are found empirically, which are equal to 0.5 and 1.0 correspondingly. Introducing the scale factor simplifies the network training since the update of thresholds is done with different learning rates for different layers of neural network as they can have various orders of values. For example, values on the intermediate layers of VGG network may increase up to 7 times in comparison with the values on the first layers.

Therefore the quantization procedure can be formalized as follows: 

\begin{equation}
T_{adj} = clip(\alpha, 0.5, 1)\cdot T_i
\end{equation}

\begin{equation}
S_I = \frac{2^n - 1}{T_{adj}}  
\end{equation}

\begin{equation}
I_q = \lfloor {I \cdot S_I} \rceil
\end{equation}

The similar procedure is performed for weights. The current quantization scheme has two non-differentiable functions, namely $round$ and $clip$. Derivatives of these functions can be defined as:

\begin{equation}
I_q = \lfloor I \rceil 
\end{equation}

\begin{equation}
\frac{dI_q}{dI} = 1
\end{equation}

\begin{equation}
X_c = clip(X, a, b)
\end{equation}

\begin{equation}
\frac{dX_c}{dX} = 
 \begin{cases}
   1, if X \in [a, b] \\    
   0, otherwise 
 \end{cases}
\end{equation}

\noindent Bias quantization is performed similar to Ref. \cite{5}:

\begin{equation}
b_q = clip(\lfloor {S_i \cdot S_w \cdot b} \rceil, -(2^{31} - 1), 2^{31} - 1)
\end{equation}

\subsubsection{Training of asymmetric thresholds.}

Quantization with symmetric thresholds described in the previous sections is easy  to implement on certain devices, however it uses an available spectrum of integer values inefficiently which significantly decreases the accuracy of quantized models. Authors in Ref. \cite{5} effectively implemented quantization with asymmetric thresholds for mobile devices, so it was decided to adapt the described above training procedure for asymmetric thresholds. 

$T_l$ and $T_r$ are left and right limits of asymmetric thresholds. However, it is more convenient to use other two values for quantization procedure: left limit and width, and train these parameters. If the left limit is equal to 0, then scaling of this value has no effect. That is why a shift for the left limit is introduced. It is calculated as:  

\begin{equation}
R = T_r - T_l 
\end{equation}

\begin{equation}
T_{adj} = T_l + clip(\alpha_T, min_{\alpha_T}, max_{\alpha_T})\cdot R
\end{equation}

The coefficients $min_{\alpha_T}$, $max_{\alpha_T}$ are set empirically. They are equal to -0.2 and 0.4 in the case of signed variables, and to 0 and 0.4 in the case of unsigned. Range width is selected in a similar way. The values of $min_{\alpha_R}$, $max_{\alpha_R}$ are also empiric and equal to 0.5 and 1.

\begin{equation}
R_{adj} = clip(\alpha_R, min_{\alpha_R}, max_{\alpha_R})\cdot R
\end{equation}

\subsubsection{Vector quantization.} 

Sometimes due to high  range of weight values it is possible to perform the discretization procedure more softly, using different thresholds for different filters of the convolutional layer. Therefore, instead of a single quantization factor for the whole convolutional layer (scalar quantization) there is a group of factors (vector quantization).  This procedure does not complicate the realization on devices, however it allows increasing the accuracy of the quantized model significantly. Considerable improvement of accuracy is observed for models with the architecture using the depth-wise separable convolutions. The most known networks of this type are MobileNet-v1 \cite{1} and MobileNet-v2 \cite{2}.

\subsection{Training on the unlabeled data}
\label{Training}

Most articles related to neural network quantization use the labeled dataset for training discretization thresholds or directly the network weights. In the proposed approach it is recommended to discard initial labels of train data which significantly speeds up transition from a trained non-quantized network to a quantized one as it reduces the requirements to the train dataset. We also suggest to optimize root-mean-square error (RMSE) between outputs of quantized and original networks before applying the softmax function, while leaving the parameters of the original network unchanged.

Suggested above technique can be considered as a special type of quantization with distillation \cite{7} where all components related to the labeled data are absent.

The total loss function $L$ is calculated by the following formula:

\begin{equation} 
\begin{split}
& L(x; W_T, W_A) = \\
& = \alpha H(y, z^T) + \beta H(y, z^A) + \gamma H(z^T, z^A)
\end{split}
\end{equation}

In our case $\alpha$ and $\beta$ are equal to 0, and 

\begin{equation}
H(z^T, z^A) = \sqrt{\sum\limits_{i=1}^N \frac{(z^T_i - z^A_i)^2}{N}}
\end{equation}

\noindent where:

\begin{itemize}
  \item $z^T$ is the output of non-quantized neural network, 
  \item $z^A$ is the output of quantized neural network, 
  \item $N$ is batch size,
  \item $y$ is the label of $x$ example.
\end{itemize}

\subsection{Quantization of depth-wise separable convolution}
\label{depth-wise}

During quantization of models having the depth-wise separable convolution layers (or DWS-layers) it was noticed that for some models (MobileNet-v2, MNasNet with the lower resolution of input images) vector quantization gives much higher accuracy than the scalar quantization. Besides, the usage of vector quantization instead of scalar only for DWS-layers gives the accuracy improvement.

In contrast to the scalar quantization, vector quantization takes into account the distribution of weights for each filter separately - each filter has its own quantization threshold. If we perform rescaling of values so that the quantization thresholds become identical for each filter, then procedures of scalar and vector quantization of the scaled data become equivalent.

For some models this approach may be inapplicable because any non-linear operations on the scaled data as well as addition of the data having different scaling factors are not allowed. Scaling the data can be made for the particular case $\mathit{DWS \rightarrow [ReLU] \rightarrow Conv}$ (see Figure 3). In this case only the weights of the model change. 

\begin{figure*}
\begin{center}
\includegraphics[width=\linewidth]{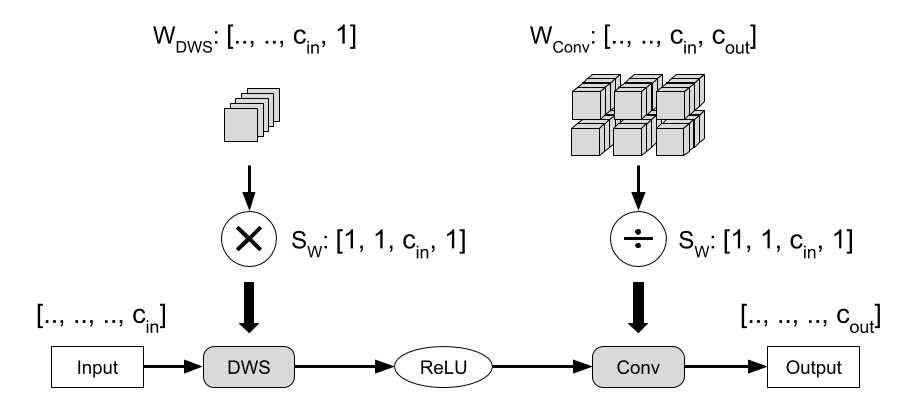}
\end{center}
\caption{Scaling the filters of DWS + Convolution layers where the output of DWS + Convolution remains unchanged. Numbers in square brackets denote the dimension of the scaling factors. $\mathit{W_{DWS}}$ represents the weights of the DWS layer, and $\mathit{W_{Conv}}$ - the weights of the convolution layer. Note that the scaling factor $S_W > 0$.}
\label{scaling}
\end{figure*}

\subsubsection{Scaling the weights for MobileNet-V2 (with ReLU6).} 

As it is mentioned above, the described method is not applicable for models which use the non-linear activation functions. In case of MobileNet, there is ReLU6 activation function between the convolutional operations. When scaling the weights of a DWS-filter, the output of the DWS-layer is also scaled. One way to keep the result of the neural network inference unchanged is to modify the ReLU6 function, so that the saturation threshold for the \textit{k-th} channel is equal to $6 \cdot S_W [k]$. However, it is not suitable for the scalar quantization technique.

In practice, the output data for some channels of a DWS-layer $X_k$ may be less than 6.0 on a large amount of input data of the network. It is possible to make rescaling for these channels, but with the certain restrictions. The scaling factor for each of these channels must be taken so that the output data for channels $X_k$ does not exceed the value 6.0.

If $X_k < 6$ and $X_k \cdot S_W [k] <6$, then

\begin{equation}
min(6, X_k \cdot S_W [k]) = S_W [k] \cdot min(6, X_k)
\end{equation}

Consequently:

\begin{equation}
ReLU6(X_k \cdot S_W [k]) = S_W [k] \cdot ReLU6(X_k)
\end{equation}

We propose the following scheme of scaling the DWS-filter weights.

\begin{enumerate} 
  \item  Find the maximum absolute value of weights for each filter of a DWS-layer. 
  \item Using the set of calibration data, determine the maximum values each channel of the output of the DWS-layer reaches (before applying ReLU6).
  \item  Mark the channels where the output values exceed 6.0 or are close to it as ``locked''. The corresponding filters of the DWS layer must stay unchanged. We propose to lock the channels where the output data is close to the value 6.0, because it could reach this value if we use a different calibration dataset. In this article we consider 5.9 as the upper limit.
  \item  Calculate the maximum absolute value of weights for each of the locked filters $T(w_i^{fixed})$. The average of these maximum values $T_0 = \overline{T(w_i^{fixed})}$ becomes a control value that is used for scaling the weights of non-locked filters. The main purpose of such choice is to minimize the difference between the thresholds of different filters of the DWS-layer.
  \item  Find the appropriate scaling factors for non-locked channels. 
  \item  Limit these scaling factors so that the maximum values on the DWS-layer output for non-locked channels do not exceed the value 6.0.
\end{enumerate}

\section{Experiments and Results}
\label{Results}
\subsection{Experiments description}
\label{Res1}
\subsubsection{Researched architectures.} 

The procedure of quantization for architectures with high redundancy is practically irrelevant because such neural networks are hardly applicable for mobile devices. Current work is focused on experiments on the architectures which are actually considered to be a standard for mobile devices (MobileNet-v2 \cite{2}), as well as on more recent ones (MNasNet \cite{3}). All architectures are tested using 224 x 224 spatial resolution.

\subsubsection{Training procedure.} 
As it is mentioned above in the section 3.2 (``Training on the unlabeled data''), we use RMSE between the original and quantized networks as a loss function. Adam optimizer \cite{20} is used for training, and cosine annealing with the reset of optimizer parameters - for learning rate. Training is carried out on approximately 10\% part of ImageNet dataset \cite{18}. Testing is done on the validation set. 100 images from the training set are used as calibration data. Training takes 6-8 epochs depending on the network.

\subsection{Results}

The quality of network quantization is represented in the Tables 1-2.  

\begin{table}
\begin{center}
\begin{tabular}{|l|c|c|c|}
\hline
Architecture   & Symmetric  & Asymmetric  & Original   \\
        	& thresholds,  & thresholds,   & accuracy, \\
            & \%     & \%      &  \%     \\
\hline\hline
MobileNet v2 & 8.1   & 19.86 & 71.55   \\         
MNas-1.0     & 72.42 & 73.46 & 74.34 \\
MNas-1.3     & 74.92 & 75.30  & 75.79 \\
\hline
\end{tabular}
\end{center}
\caption{Quantization in the 8-bit scalar mode.}
\label{table1}
\end{table}

Experimental results show that the scalar quantization of MobileNet-v2 has very poor accuracy. A possible reason of such quality degradation is the usage of ReLU6 activation function in the full-precision network. Negative influence of this function on the process of network quantization is mentioned in Ref. \cite{21}. In case of using vector procedure of thresholds calculation, the accuracy of quantized MobileNet-v2 network and other researched neural networks is almost the same as the original one.

The Tensorflow framework \cite{9} is chosen for implementation because it is rather flexible and convenient for further porting to mobile devices.
Pre-trained networks are taken from Tensorflow repository \cite{24}. To verify the results, the program code and quantized scalar models in the .lite format, ready to run on mobile phones, are presented in the repository specified in the abstract.

\begin{table}
\begin{center}
\begin{tabular}{|l|c|c|c|}
\hline
Architecture   & Symmetric  & Asymmetric  & Original   \\
        	& thresholds,  & thresholds,   & accuracy, \\
            & \%     & \%      &  \%     \\
\hline\hline
MobileNet v2  & 71.11  & 71.39 & 71.55 \\
MNas-1.0     & 73.96 & 74.25 & 74.34 \\
MNas-1.3     & 75.56 & 75.72 & 75.79 \\
\hline
\end{tabular}
\end{center}
\caption{Quantization in the 8-bit vector mode.}
\label{table2}
\end{table}

The algorithm described in the section 3.3 (``Quantization of depth-wise separable convolution'') gives the following results. After performing the scalar quantization of the original MobileNetV2 model, its accuracy becomes low (the top-1 value is about 1.6\%). Applying the weights rescaling before the quantization increases the accuracy of the quantized model up to 67\% (the accuracy of the original model is 71.55\% \footnote{The network accuracy is measured on a full validation set ImageNet2012 which includes single-channel images.}). To improve the accuracy of the quantized model we use fine-tuning of weights for all filters and biases. Fine-tuning is implemented via trainable point-wise scale factors where each value can vary from 0.75 to 1.25. The intuition behind this approach is to compensate the disadvantages of the linear quantization by slight modification of weights and biases, so some values can change their quantized state. As a result, fine-tuning improves the accuracy of the quantized model up to 71\% (without training the quantization thresholds). Fine-tuning procedures are the same as described in the section 4.1 (``Experiments description'').

\section{Conclusion}
\label{Conclusion}

This paper demonstrates the methodology of neural network quantization with fine-tuning. Quantized networks obtained with the help of our method demonstrate a high accuracy that is proved experimentally. Our work shows that setting a quantization threshold as multiplication of the maximum threshold value and trained scaling factor, and also training on a small set of unlabeled data allow using the described method of quantization for fast conversion of pre-trained models to mobile devices.

\end{document}